\documentclass[onecolumn,12pt]{IEEEtran}
\usepackage{cite}
\usepackage{amsmath,amssymb,amsfonts,amsthm}
\usepackage{algorithm}
\usepackage{algpseudocode}
\usepackage{graphicx}
\usepackage{textcomp}
\usepackage{xcolor}
\usepackage[colorlinks=true, linkcolor=blue, citecolor=blue, urlcolor=blue]{hyperref}
\usepackage{tikz}
\usetikzlibrary{decorations.pathreplacing, positioning}

\begin{document}
\title{HRA: A Multi-Criteria Framework for Ranking Metaheuristic Optimization Algorithms}
\author{\IEEEauthorblockN{Evgenia-Maria~K.~Goula  and  Dimitris~G.~Sotiropoulos}\\
        \IEEEauthorblockA{\textit{Department of Electrical and Computer Engineering} \\
                          \textit{School of Engineering, University of Peloponnese} \\ 
                          GR 263 34, Patras, Greece \\
                               Email:   \{e.goula, dg.sotiropoulos\}@uop.gr}
        }

\maketitle

\thispagestyle{empty} 

\begin{abstract}
Metaheuristic algorithms are essential for solving complex optimization problems in different fields. 
However, the difficulty in comparing and rating these algorithms remains due to the wide range of performance metrics and problem dimensions usually involved. On the other hand, nonparametric statistical methods and post hoc tests are time-consuming, especially when we only need to identify the top performers among many algorithms. The Hierarchical Rank Aggregation (HRA) algorithm aims to efficiently rank metaheuristic algorithms based on their performance across many criteria and dimensions. The HRA employs a hierarchical framework that begins with collecting performance metrics on various benchmark functions and dimensions. Rank-based normalization is employed for each performance measure to ensure comparability and the robust TOPSIS aggregation is applied to combine these rankings at several hierarchical levels, resulting in a comprehensive ranking of the algorithms.
Our study uses data from the CEC 2017 competition to demonstrate the robustness and efficacy of the HRA framework. It examines 30 benchmark functions and evaluates the performance of 13 metaheuristic algorithms across five performance indicators in four distinct dimensions. This presentation highlights the potential of the HRA to enhance the interpretation of the comparative advantages and disadvantages of various algorithms by simplifying practitioners' choices of the most appropriate algorithm for certain optimization problems.

\end{abstract}
\begin{IEEEkeywords}
Multi-Criteria Decision Making (MCDM), Metaheuristic algorithms, Hierarchical Rank Aggregation, Rank-Based Normalization, Robust TOPSIS, CEC2017
\end{IEEEkeywords}

\section{Introduction}\label{sec:intro}
Computational intelligence has emerged as a rapidly evolving field, with metaheuristic algorithms playing a crucial role in solving complex optimization problems in various domains \cite{mirjalili_2023}. These algorithms, inspired by diverse sources such as natural evolutionary processes, swarm behavior, and mathematical constructs, have successfully tackled challenging optimization tasks \cite{Ma_2023}. 
However, the No Free Lunch, or NFL Theorem, states that no algorithm should consistently outperform all others across all problem domains \cite{Adam_2019}. 
This fundamental principle, coupled with the ever-growing number of metaheuristic algorithms-now exceeding 500 plus \cite{Ma_2023}, underscores the critical need for robust and comprehensive assessment methods to compare and evaluate algorithm performance.

Traditionally, algorithm performance has been assessed using statistical and non-parametric techniques. Though applicable, such evaluations are very much restricted in that they concentrate only on mean performance measures and would probably neglect other essential characteristics of how an algorithm works. Pacheco et al. {\cite{Pacheco_2018}} presented the limitations of such approaches and emphasized the fact that such procedures primarily neglect the measures of dispersion, and in particular, the standard deviation of the obtained results, which is very crucial when analyzing performance and reliability of algorithms.
This concern implies that researchers and practitioners who desire to utilize optimization methods or work on the improvement of algorithm designs may, therefore, tend to underestimate a certain algorithm in its capability and applicability.
However, in the last decade, more researchers have started employing a systematic method for the problem of algorithm performance by utilizing Multi-Criteria Decision-Making (MCDM) techniques within algorithm evaluation based on multiple often conflicting criteria. 

The developments in MCDM and its applications in evaluating the algorithms can be highlighted in several contributions. In an early work by Peng et al. \cite{Peng_2011}, MCDM methods were combined to rank and compare classification algorithms, among very few attempts of multi-criteria evaluation in machine learning. Kou et al. \cite{KOU_2012} extended this work by proposing a framework based on Spearman’s rank correlation coefficient to resolve conflicting outcomes using multiple MCDM methods. Later, in 2014, Kou et al. applied an MCDM-based approach to rank popular clustering algorithms in financial risk analysis, demonstrating the effectiveness of three MCDM methods across six clustering algorithms and multiple real-life datasets \cite{Kou_2014}.

A significant contribution in employing MCDM techniques to evaluate metaheuristic algorithms can be traced to 2015 when Krohling and Pacheco introduced A-TOPSIS \cite{Krohling_2015a}. This new approach, similar to the Technique for Order Preference by Similarity to Ideal Solution (TOPSIS) has been created to fully resolve the issue of evolutionary algorithm comparison. A-TOPSIS advances the use of mean and standard deviation in the evaluation process, which improves the conventional statistical tests and ranking strategies, particularly the Friedman test that only considers means for ranking. Krohling et al. \cite{Krohling_2015b} made further developments in this area with Hellinger-TOPSIS which employs the Hellinger distance for distance measurement, within which it is possible to differentiate the performance of algorithms under comparison. 

Yu et al.\cite{Yu_2018} presented for the first time a new framework for assessing the performance of multiobjective evolutionary algorithms, 
which was the first to demonstrate MCDM techniques for complex problems with multiple performance measures. In the same direction,  Barak and Mokfi \cite{Barak_2019} developed an MCDM-based model focused on assessing clustering algorithms, where group MCDM techniques and the Borda count method are employed. This suggests new approaches to evaluating algorithms in the fields of evolutionary computation and data clustering.
The adoption of MCDM for assessing meta-heuristic algorithms as well as their components continues to develop; for instance, Shadkam et al. \cite{Shadkam_2021} provide insight on the ranking of metaheuristic algorithms using AHP, TOPSIS, and AHP-TOPSIS combining techniques, while  Balaji et al. \cite{L_2021}, employed the TOPSIS method to rank meta-heuristic algorithms with a consideration of groundwater vulnerability assessment as the application context. Tabassum and Akram \cite{tabassum_2021} introduced a Rank-based TOPSIS (RB-TOPSIS)  approach for evaluating the performance of metaheuristic algorithms, contributing significantly to the field of MCDM in algorithm assessment in their study applied CEC2017 benchmark. 

The area of metaheuristic optimization has several limitations concerning algorithm evaluation and comparison, some more critical than others. These include the inability to combine multiple performance metrics, the need to evaluate performance on different problem dimensions, and the lack of clarity regarding providing an overall ranking. Moreover, the rapid increase in the number of algorithms and diversity in the optimization problems calls for more rigorous evaluation strategies.

The Hierarchical Rank Aggregation (HRA) framework presented in this paper addresses these challenges by providing a systematic and multi-granular perspective of how metaheuristic optimization algorithms should be evaluated or ranked. The HRA utilizes the robust TOPSIS technique in a hierarchical structure, making it possible to assess performance measures from different problem dimensions. This assessment approach increases the correctness of the relationships among algorithms and makes it easier to understand the algorithm performance on different levels of a given problem. To illustrate how our proposed framework works, we utilized it to evaluate 13  metaheuristic algorithms against 30 benchmark functions as part of the CEC 2017 competition dataset in four dimensions. Based on this analysis, we intend to hand over to the computational intelligence community a robust algorithm assessment methodology, enabling more rational algorithm selection and development processes in the optimization domain.

 The remainder of this paper is organized as follows: Section~\ref{sec:methodology} explains the components of the HRA framework in detail step by step, incorporating the R-TOPSIS procedural structure and hierarchical aggregation. The usage of the HRA algorithm in the CEC'2017 competition dataset is demonstrated in Section~\ref{sec:Results}, and a detailed discussion of the results was presented. The Section~~\ref{sec:conclusion} summarizes present research and describes the plans for the next one.

\section{Methodology}\label{sec:methodology}
This research, in particular, aims to develop an exhaustive ranking system for the metaheuristic algorithms tested to consider how successfully a particular algorithm performs across various benchmark functions and problem dimensions. Let us define: \\
$\mathcal{A} = \{ A_1, \ldots, A_m \}$ as the set of algorithms, \\
$\mathcal{F} = \{ f_1, \ldots, f_n \}$ as the set of benchmark functions,\\
$\mathcal{D} = \{ d_1, \ldots, d_k \}$ as the set of problem dimensions, and\\
$\mathcal{P} = \{ p_1, \ldots, p_l \}$ as the set of performance measures.\\
For each dimension $d \in \mathcal{D}$, performance measure $p \in \mathcal{P}$, algorithm $A \in \mathcal{A}$, and benchmark function $f \in \mathcal{F}$, we have the performance values $X_{d,p,A,f}$ of algorithm $A$ on benchmark function $f$ under dimension $d$ for performance measure $p$. These values form a decision matrix $m\times n$.

Our approach involves converting raw performance data into ranks within each criterion of a decision matrix and then creating a rank-based decision matrix. This approach is helpful since relative performance is more critical than absolute value. It also provides an additional advantage of avoiding scale effects across evaluation criteria and biasing from the outlier scores since ranks are used instead of raw scores.

The ranked data are then processed using the R-TOPSIS method described by Aires and Ferreira \cite{Aires_2019}. 
This method normalizes the performance measures and then ranks the evaluated measures using a hierarchical structure, as shown in Fig.~\ref{fig:hierarchical_structure}, by ensuring robust and interpretable rankings reflecting algorithm performance's multifaceted nature across different benchmarks.

\subsection{The R-TOPSIS Method}

TOPSIS (Technique for Order Preference by Similarity to Ideal Solution) is a multi-criteria decision-making method (MCDM) used to select the best alternative from a set of options based on multiple criteria. It was initially proposed by Hwang and Yoon in 1981 and is widely used in decision analysis and operations research  \cite{Hwang_1981}. TOPSIS is a straightforward ranking method both in concept and in application. The core principle of the standard TOPSIS method is to select alternatives with the shortest distance from the positive ideal solution (PIS) and the farthest distance from the negative ideal solution (NIS). The ideal positive solution emphasizes maximizing benefit criteria and minimizing cost criteria, while the negative solution focuses on maximizing and minimizing cost criteria. However, TOPSIS has faced criticism for its rank reversal issue. Rank reversal is when the order of alternatives changes after adding or removing an alternative from a previously rated group.

Aires and Ferreira \cite{Aires_2019} introduced R-TOPSIS as a solution to address the issue in TOPSIS. To this end, the authors introduced an additional input parameter, the domain, to describe the range of possible values for each criterion. 
Furthermore, the type of normalization adopted, whether Max-Min or Max normalization, helped to maintain optimal solutions and the stability of the normalized and weighted decision matrices, even in case some adjustments were made to the original decision problem. These improvements focused on increasing the reliability and logic of TOPSIS while maintaining its user-friendliness. The steps that will be used to apply the R-TOPSIS method have been outlined below.

\medskip\noindent
\emph{Step 1:} \  Define a set of alternatives $A = \{a_i\}_{i=1}^m$ \\
\medskip\noindent
\emph{Step 2:} \ Define a set of criteria $C = \{c_j\}_{j=1}^n$ and a subdomain 
$D = \{d_j\}_{j=1}^n$, where $d_{1j}$ is the minimum value and $d_{2j}$ is the maximum value of $D_j$ \\
\medskip\noindent
\emph{Step 3:} \  Estimate the performance rating of the alternatives as $X = \{x_{ij}\}_{i=1,j=1}^{m,n}$ \\
\medskip\noindent
\emph{Step 4:} \ Elicit the criteria weights as $W = \{w_j\}_{j=1}^n$, where $w_j > 0$ and $\sum_{j=1}^n w_j = 1$ \\
\medskip\noindent
\emph{Step 5:} \  Calculate the normalized decision matrix $(n_{ij})$ \\
\medskip\noindent
\phantom{}
\hspace{1em} \emph{Step 5.1:} \ Max  normalization
        \[
        n_{ij} = \frac{x_{ij}}{d_{2j}}, \quad i = 1, 2, \ldots, m; \quad j = 1, 2, \ldots, n
        \]
\phantom{}\hspace{1em} \emph{Step 5.2:} \ Max-Min normalization 
        \[
        n_{ij} = \frac{x_{ij} - d_{1j}}{d_{2j} - d_{1j}}, \quad i = 1, 2, \ldots, m; \quad j = 1, 2, \ldots, n
        \]
\noindent
\emph{Step 6:} \ Calculate the weighted normalized decision matrix $(r_{ij})$ as:
        \[
        r_{ij} = w_j \times n_{ij}, \quad i = 1, 2, \ldots, m; \quad j = 1, 2, \ldots, n
        \]
\noindent
\emph{Step 7:} \ Set the negative (NIS) and positive (PIS) ideal solutions
        \[
        NIS = \{r_j^{-}\}_{j=1}^n, \quad \text{where } r_j^{-} = \begin{cases} 
        \frac{d_{1j}}{d_{2j}} w_j & \text{if } j \in \text{Benefit} \\ 
        w_j & \text{if } j \in \text{Cost} 
        \end{cases}
        \]
        \[
        PIS = \{r_j^{+}\}_{j=1}^n, \quad \text{where } r_j^{+} = \begin{cases} 
        w_j & \text{if } j \in \text{Benefit} \\ 
        \frac{d_{1j}}{d_{2j}} w_j & \text{if } j \in \text{Cost} 
        \end{cases}
        \]
\noindent
\emph{Step 8:} \ Calculate the distances of each alternative $i$ in relation to the ideal solutions
        \[
        S_i^{+} = \sqrt{\sum_{j=1}^n (r_{ij} - r_j^{+})^2}, \quad i = 1, 2, \ldots, m
        \]
        \[
        S_i^{-} = \sqrt{\sum_{j=1}^n (r_{ij} - r_j^{-})^2}, \quad i = 1, 2, \ldots, m
        \]
\noindent
\emph{Step 9:} Calculate the closeness coefficient of the alternatives $(CC_i)$ as:
        \[
        CC_i = \frac{S_i^{-}}{S_i^{+} + S_i^{-}}, \quad i = 1, 2, \ldots, m
        \]
\noindent
\emph{Step 10:} Sort the alternatives in descending order. The highest $CC_i$ value indicates the best performance concerning the evaluation criteria. Return sorted alternatives based on $CC_i$.

\medskip
In applying the R-TOPSIS method, we impose the domain as $[0, m+1]^n$, $m$ being the number of algorithms, while $n$ is the number of criteria. A significant benefit of this specified domain is that the Max and Max-Min normalization techniques used in Step~5 of R-TOPSIS are equivalent, and there is no reason to adopt either method and provide the rationale. This simplifies the decision-making process when utilizing R-TOPSIS because it guarantees uniform and dependable normalization of performance indices across various criteria.

\subsection{HRA: Detailed Algorithmic Steps}

The following outlines the basic steps of the Hierarchical Rank Aggregation (HRA) algorithm, which ensures a clear assessment and robust ranking of alternatives.

\medskip\noindent
\emph{1)} \  Collect the performance measures $ X_{d,p,A,f} $, which constitute $m \times n$ matrices for each combination of dimensions $d$ and performance measures $p$, resulting in a total of $l\cdot k$ matrices. Each matrix has $m$ rows corresponding to the algorithms and $n$ columns corresponding to the benchmark functions. The general form of these matrices is given by:
$$
X_{d,p} = \begin{bmatrix}
X_{d,p,A_1,f_1} & X_{d,p,A_1,f_2} & \cdots & X_{d,p,A_1,f_n} \\
X_{d,p,A_2,f_1} & X_{d,p,A_2,f_2} & \cdots & X_{d,p,A_2,f_n} \\
\vdots          & \vdots          & \ddots & \vdots           \\
X_{d,p,A_m,f_1} & X_{d,p,A_m,f_2} & \cdots & X_{d,p,A_m,f_n}
\end{bmatrix}
$$
where each element $ X_{d,p,A_i,f_j} $ represents the performance value of algorithm $ A_i $ on benchmark function $ f_j $ under dimension $d$ for performance measure $p$. All benchmark functions are considered equivalent and assign an equal weight of $1/n$.

\medskip\noindent
\emph{2)} \ Apply the mean rank transformation on each column of the matrices $ X_{d,p}$ ensuring comparability and robustness during the application of RTOPSIS. The general form of the resulting ranked matrices $ R_{d,p} $ is given by:
$$
R_{d,p} = \begin{bmatrix}
R_{d,p,A_1,f_1} & R_{d,p,A_1,f_2} & \cdots & R_{d,p,A_1,f_n} \\
R_{d,p,A_2,f_1} & R_{d,p,A_2,f_2} & \cdots & R_{d,p,A_2,f_n} \\
\vdots          & \vdots          & \ddots & \vdots           \\
R_{d,p,A_m,f_1} & R_{d,p,A_m,f_2} & \cdots & R_{d,p,A_m,f_n}
\end{bmatrix}
$$
where each element $R_{d,p,A_i,f_j}$ represents the rank of algorithm $A_i$ on benchmark function $f_j$ under dimension $d$ for performance measure $p$.
All these matrices $ R_{d,p} $ can be visualized as the leaves of a tree, constituting the fundamental component of a hierarchical framework.

\begin{figure*}[t]
\centering
\resizebox{1.0\textwidth}{!}{%
\begin{tikzpicture}[node distance=0.5cm and 1cm]
  \node (Rdp1) [rectangle, draw] {$R_{d_1,p_1}$};
  \node (Rdp2) [rectangle, draw, right of=Rdp1, xshift=1cm] {$R_{d_1,p_2}$};
  \node (dots1) [right of=Rdp2, xshift=0.5cm] {$\ldots$};
  \node (Rdp3) [rectangle, draw, right of=dots1, xshift=0.5cm] {$R_{d_1,p_\ell}$};
  
  \node (Cd1) [rectangle, draw, above of=dots1, xshift=-0.8cm, yshift=1.5cm] {$C_{d_1}$};
  \node (dimC1) [left of=Cd1, xshift=-0.5cm, scale=0.75] {$m \times \ell$};
  
  \node (Rdp4) [rectangle, draw, right of=Rdp3, xshift=1cm] {$R_{d_2,p_1}$};
  \node (Rdp5) [rectangle, draw, right of=Rdp4, xshift=1cm] {$R_{d_2,p_2}$};
  \node (dots2) [right of=Rdp5, xshift=0.5cm] {$\ldots$};
  \node (Rdp6) [rectangle, draw, right of=dots2, xshift=0.5cm] {$R_{d_2,p_\ell}$};
  
  \node (Cd2) [rectangle, draw, above of=dots2, xshift=-0.8cm, yshift=1.5cm] {$C_{d_2}$};
  \node (dimC2) [left of=Cd2, xshift=-0.5cm, scale=0.75] {$m \times \ell$};

  \node (dots4) [right of=Rdp6, xshift=0.5cm] {$\ldots$};
  
  \node (Rdp7) [rectangle, draw, right of=dots4, xshift=0.5cm] {$R_{d_k,p_1}$};
  \node (Rdp8) [rectangle, draw, right of=Rdp7, xshift=1cm] {$R_{d_k,p_2}$};
  \node (dots3) [right of=Rdp8, xshift=0.5cm] {$\ldots$};
  \node (Rdp9) [rectangle, draw, right of=dots3, xshift=0.5cm] {$R_{d_k,p_\ell}$};
  
  \node (Cd3) [rectangle, draw, above of=dots3, xshift=-0.8cm, yshift=1.5cm] {$C_{d_k}$};
  \node (dimC3) [left of=Cd3, xshift=-0.5cm, scale=0.75] {$m \times \ell$};
  
  \node (dots5) [right of=Cd2, xshift=1.5cm] {$\ldots$};
  
  \node (Cfinal) [rectangle, draw, above of=Cd2, yshift=1.5cm]{$D$};
  \node (dimCinal) [left of=Cfinal, xshift=-0.5cm, scale=0.75] {$m \times k$};
  
  \draw[->] (Rdp1) -- node[midway, sloped, above, scale=0.5] {RTOPSIS} (Cd1);
  \draw[->] (Rdp2) -- node[midway, sloped, above, scale=0.5] {RTOPSIS} (Cd1);
  \draw[->] (Rdp3) -- node[midway, sloped, above, scale=0.5] {RTOPSIS} (Cd1);
  
  \draw[->] (Rdp4) -- node[midway, sloped, above, scale=0.5] {RTOPSIS} (Cd2);
  \draw[->] (Rdp5) -- node[midway, sloped, above, scale=0.5] {RTOPSIS} (Cd2);
  \draw[->] (Rdp6) -- node[midway, sloped, above, scale=0.5] {RTOPSIS} (Cd2);

  \draw[->] (Rdp7) -- node[midway, sloped, above, scale=0.5] {RTOPSIS} (Cd3);
  \draw[->] (Rdp8) -- node[midway, sloped, above, scale=0.5] {RTOPSIS} (Cd3);
  \draw[->] (Rdp9) -- node[midway, sloped, above, scale=0.5] {RTOPSIS} (Cd3);
    
  \draw[->] (Cd1) -- node[midway, sloped, above, scale=0.5] {RTOPSIS} node[midway, sloped, below, scale=0.75] {$C_{d_1}^{*}$} (Cfinal);
  \draw[->] (Cd2) -- node[midway, sloped, above, scale=0.5] {RTOPSIS} node[midway, sloped, below, scale=0.75] {$C_{d_2}^{*}$} (Cfinal);
  \draw[->] (Cd3) -- node[midway, sloped, above, scale=0.5] {RTOPSIS} node[midway, sloped, below, scale=0.75] {$C_{d_k}^{*}$} (Cfinal);
  
  \draw[decorate,decoration={brace,amplitude=10pt,mirror},yshift=1.8cm] (Rdp1.south west) -- (Rdp3.south east) node[midway,below,yshift=-0.5cm,scale=0.75]{$\ell$ matrices $m \times n$};
  
  \draw[decorate,decoration={brace,amplitude=10pt,mirror},yshift=1.8cm] (Rdp4.south west) -- (Rdp6.south east) node[midway,below,yshift=-0.5cm,scale=0.75]{$\ell$ matrices $m \times n$};
  
  \draw[decorate,decoration={brace,amplitude=10pt,mirror},yshift=1.8cm] (Rdp7.south west) -- (Rdp9.south east) node[midway,below,yshift=-0.5cm, scale=0.75]{$\ell$ matrices $m \times n$};
  
\end{tikzpicture}
}%
\caption{Tree structure of the Hierarchical Rank Aggregation (HRA) algorithm using RTOPSIS.}
\label{fig:hierarchical_structure}
\end{figure*}
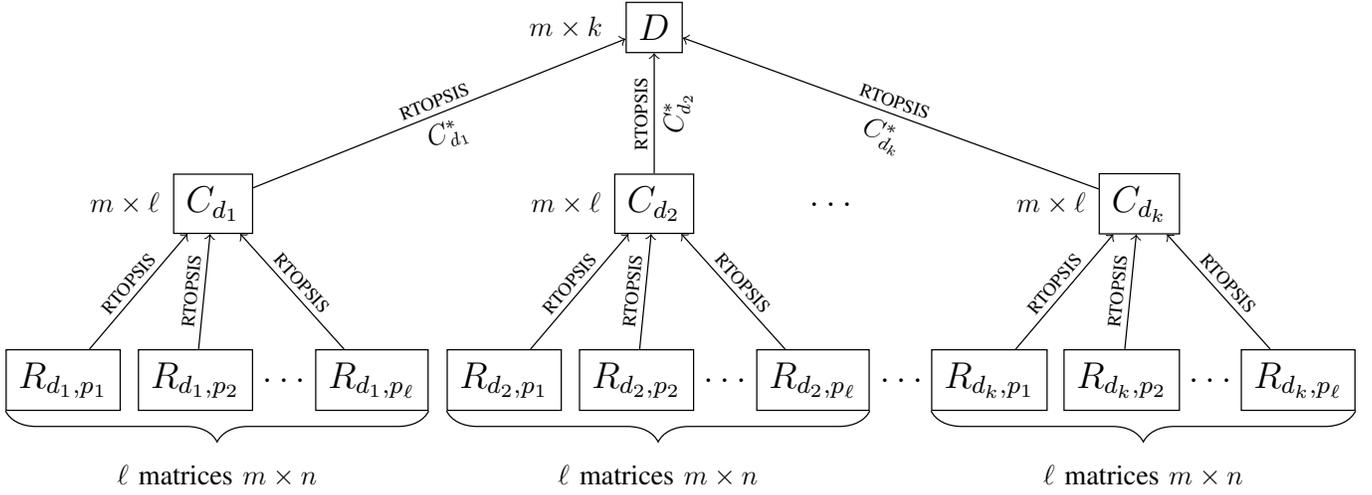

\medskip\noindent
\emph{3)} \ Apply the RTOPSIS method to the rank decision matrices $ R_{d,p} $ (of size $ m \times n $) for the $ d $-th dimension and $ p $-th performance measure:
$
C_{d,p} = \text{RTOPSIS}(R_{d,p}, w_{\mathcal{F}})
$
where ${C}_{d,p}$ is the resulting $m\times 1$ vector of ranks obtained. The weight vector ${w}_F$ assigns equal importance to all the benchmark functions, with each weight being $1/n$ for $j = 1,\ldots, n$.  The vector ${C}_{d,p}$ provides the aggregated ranks of the algorithms for the specified dimension and performance measure, indicating their relative performance. This information will be synthesized at the next level of the hierarchy by grouping the performance measures for each dimension (see Fig.~\ref{fig:hierarchical_structure}). 

\smallskip\noindent
\emph{4)} \ Once the rank vectors $C_{d,p}$ have been derived via the TOPSIS method for each combination of dimension $d$ and performance measure $p$, these vectors are concatenated at the second level of the hierarchical structure to create the intermediate matrices $C_d$. Each intermediate matrix $C_{d} = \begin{bmatrix} C_{d,p_1} & C_{d,p_2} & \cdots & C_{d,p_\ell}   \end{bmatrix}$, encapsulates the performance ranks for a particular dimension across all performance measures. 
Therefore, at this level of the hierarchical structure (see Fig.~\ref{fig:hierarchical_structure}), we have a total of $k$ in numbers, $C_{d}$ matrices, corresponding to the number of function dimensions. Consequently, we may examine the performance of the algorithms for each dimension, considering the performance measures evaluated $\ell$.

\smallskip\noindent
\emph{5)} \ Apply the RTOPSIS method on the intermediate matrices $C_{d}$ for each dimension $d$ to determine the definitive ranking of algorithms per dimension. This procedure integrates performance metrics to generate a unified ranking for each dimension:
$
C_d^{*} = \text{RTOPSIS}(C_{d}, w_{\mathcal{P}}), 
$
where $w_{\mathcal{P}}$ the corresponding weight vector.
The outcome $C_d^{*}$ is a vector $m\times 1$, which presents the overall ranking of the $m$ algorithms for dimension $d \in \mathcal{D}$ incorporating all performance metrics for the specified dimension.

\smallskip\noindent
\emph{6)} \  The final stage involves the construction of the decision matrix $D$ of size $m \times k$, which constitutes from  the rank vectors $C_d^{*}$ as 
$
D = \begin{bmatrix} 
    C_{d_{1}}^{*} & C_{d_{2}}^{*} & \cdots & C_{d_{k}}^{*}
    \end{bmatrix}
$
and apply the TOPSIS method one final time. 
The overall ranking vector $D^{*}$ is obtained by:
$
D^{*} = \text{RTOPSIS}(D, w_{\mathcal{D}}).
$
The weight vector $w_{\mathcal{D}}$ determines the relative importance of each dimension.

This thorough rating enables a full assessment of the algorithms' efficacy, reflecting their performance in a multidimensional framework.

\subsection{Tree Structure of HRA Algorithm}

This process can further be understood from a hierarchical perspective, which is a hierarchical tree in nature, if we assume that we can design the RTOPSIS method within this framework. Each tissue of leaf of the graph is a rank decision matrix $R_{d,p}$, while edges bearing the title ``RTOPSIS'' show the RTOPSIS application. These edges provide information to form an intermediate matrix $C_{d}$ of size $m \times \ell$ for each dimension $d$. Out of total of $k$, there are such matrices pertaining to each dimension. The ranks aggregate that pour all these matrices together are used to determine the decision matrix for the ultimate ranking of the algorithms.

We can visualize this process as a tree to better understand the hierarchical structure and the application of the RTOPSIS method. Each leaf node represents a rank decision matrix $R_{d,p}$, and the edges, labeled ``RTOPSIS'', indicate the application of the RTOPSIS method. These edges provide information to form an intermediate matrix $C_{d}$ of size $m \times \ell$ for each dimension $d$. We have a total of $k$ such matrices, one for each dimension. The aggregated ranks of these matrices are then combined to form the final decision matrix for the overall ranking of the algorithms.

Fig.~\ref{fig:hierarchical_structure} elucidates this process clearly and structured. Each rank decision matrix $R_{d,p}$ (where $d$ represents the dimension and $p$ the particular criterion within that dimension) undergoes the RTOPSIS transformation to produce a corresponding ranking vector $C_{d,p}$. These vectors, which represent the performance scores of the algorithms for each criterion, are subsequently consolidated into an intermediate matrix $C_{d}$. This intermediate matrix $C_{d}$ has dimensions $m \times \ell$, where $m$ is the number of algorithms, and $\ell$ is the number of criteria within dimension $d$.

The process is repeated for all $k$ dimensions, resulting in intermediate $k$ matrices $C_{d}$. Each of these intermediate matrices $C_{d}$ is then aggregated into a single decision matrix $D$ of size $m \times k$. This final decision matrix $D$ encapsulates the performance of the algorithms across all dimensions.

In the final step, we reapply the RTOPSIS method to the decision matrix $D$ to obtain the overall ranking vector $D^{*}$. This vector $D^{*}$ provides a comprehensive ranking of the algorithms, reflecting their performance across all considered dimensions and criteria.

Thus, as shown in Fig.~\ref{fig:hierarchical_structure}, the hierarchical structure ensures a systematic and thorough evaluation of the algorithms. It allows for the incorporation of multiple dimensions and criteria, facilitating a multidimensional assessment and yielding a final ranking that is both holistic and robust.

\begin{algorithm}[tb]
\caption{Hierarchical Rank Aggregation (HRA)}\label{alg:HRA}
\begin{algorithmic}[1]
\Require Set of algorithms $\mathcal{A} = \{ A_1, \ldots, A_m \}$, set of benchmark functions $\mathcal{F} = \{ f_1, \ldots, f_n \}$, set of dimensions $\mathcal{D} = \{ d_1, \ldots, d_k \}$, set of performance measures $\mathcal{P} = \{ p_1, \ldots, p_l \}$, performance values $X_{d,p,A,f}$
\Ensure Overall ranking vector $D^*$

\For{each dimension $ d \in \mathcal{D} $}
    \For{each performance measure $ p \in \mathcal{P} $}
        \State Collect performance metrics $X_{d,p}$
        \State Rank them to obtain $R_{d,p}$
        \State Apply RTOPSIS to $R_{d,p}$ to get $C_{d,p}$
    \EndFor
    \State Form intermediate matrix $C_d$ from $C_{d,p}$ vectors
   \State Apply RTOPSIS to $C_d$ to get $C_d^*$
\EndFor
\State Construct decision matrix $D$ from $C_d^*$ vectors
\State Apply RTOPSIS to $D$ to get overall ranking $D^*$
\State \Return $D^*$
\end{algorithmic}
\end{algorithm}

When applying the RTOPSIS method in HRA, the weights must be assigned to different criteria, represented as $w_{\mathcal{F}}$, $w_{\mathcal{P}}$, and $w_{\mathcal{D}}$. Although the precise determination of these weights can substantially impact the outcomes, we will not explore the intricacies of this issue in the present work. One rough method for estimating these weights is to give each criterion equal significance, thereby employing equal weights. 

However, this approach could result in a significant loss of information if it does not consider the relative importance of each criterion.
In practice, it is reasonable to prioritize the criteria by ranking them, as rankings are often simpler to provide than precise numerical weights, which can be challenging to quantify.  One possible solution to this problem is the surrogate weight approach, commonly used to determine weights based on ordinal ranking \cite{Barron_1996, Danielson_2016}. This method may improve the prescription of solving several criteria problems in evaluating algorithms, such as ensuring the inclusion of expert opinion into the decision-making process and not needing exact numerical values. 

It is worth noticing that the HRA algorithm exhibits remarkable computational efficiency; we can easily prove that it achieves an overall time complexity of $\mathcal{O}(m \log m)$, where $m$ denotes the number of algorithms under comparison. Moreover, it is characterized by 
a problem-dependent constant $c = 1 + k + \ell k$, where $k$ represents the number of dimensions and $\ell$ is the number of performance measures. This efficiency is advantageous compared to pairwise comparison methods $\mathcal{O}(m^2)$.
The tree-structured design of HRA not only supports this time complexity but also allows potential parallelization at each level, making it a practical tool for metaheuristic algorithm comparison and ranking, especially when dealing with a large number of algorithms or extensive datasets.

\subsection{The CEC'17 Competition Dataset}
The CEC'17 test suite, featured in the 2017 IEEE Congress on Evolutionary Computation, comprises 30 distinct optimization problems (denoted as $f_1-f_{30}$) categorized into four groups: 3 unimodal, 7 multimodal, 10 hybrid, and 10 composition functions. The competition's objective was to minimize these test functions $f(x)$, where $x \in \mathbb{R}^d$ and $d \in \{10, 30, 50, 100\}$. Detailed descriptions of these functions are provided in the technical report by Awad et al. \cite{CEC2017}.
Thirteen metaheuristic algorithms were evaluated in the CEC'17 competition: jSO \cite{Brest_2017}, MM-OED \cite{sallam_2017}, IDEbestNsize \cite{Bujok_2017}, RB-IPOP-CMA-ES \cite{Biedrzycki_2017}, LSHADE-SPACMA \cite{Mohamed_2017}, DES \cite{Jagodzinski_2017}, DYYPO \cite{Maharana_2017}, TLBO-FL \cite{Kommadath_2017}, PPSO \cite{Tangherloni_2017}, MOS-SOCO2011 \cite{LaTorre_2017}, MOS-SOCO2013 \cite{LaTorre_2017}, LSHADE-cnEpSin \cite{Awad_2017}, and EBOwithCMAR \cite{Kumar_2017}. Each algorithm was run 51 times independently for each function and dimension.
To ensure the reliability and reproducibility of our analysis, we collected the raw data directly from the competition organizer's GitHub repository\footnote{\url{https://github.com/P-N-Suganthan/CEC2017-BoundContrained}}.

This dataset includes the function error values $(f_i(x) - f_i(x^*))$ for each algorithm, function, and dimension.
We developed a series of MATLAB scripts to compute and organize the statistical measures for our analysis using these raw data; the scripts are available in our GitHub repository\footnote{\url{https://github.com/dgsotiropoulos/CEC-Benchmarks}}, process the raw data to calculate five key statistical measures: best, worst, median, mean, and standard deviation of the error values.
The processed data are stored in four Excel files, one for each problem dimension:
\begin{itemize}
    \item \texttt{CEC2017\_Dim\_10.xlsx}, 
    \item \texttt{CEC2017\_Dim\_30.xlsx},
    \item \texttt{CEC2017\_Dim\_50.xlsx}, and 
    \item \texttt{CEC2017\_Dim\_100.xlsx} 
\end{itemize}
Each Excel file contains five sheets, corresponding to the statistical measures $\mathcal{P} = \{\text{best}, \text{worst}, \text{median}, \text{mean}, \text{std}\}$. Within each sheet, we store a $30 \times 13$ table, where rows represent the 30 test functions $\mathcal{F} = \{f_1, \ldots, f_{30}\}$ and columns represent the 13 algorithms. 

This structured approach to data collection and processing not only provides a sound and thorough basis for our analysis, which is expected to follow but also helps meet the requirements of the hierarchy of the HRA framework we are proposing. Such a way of compiling the data corresponds well with the multi-level approach to HRA since we create performance measures-focused and dimension-based Excel files with several sheets for different performance measures. Further, this particular organization of the data makes the execution of our algorithm quite straightforward, such that RTOPSIS can be more efficiently called at all levels of the hierarchy.

\section{Results and Discussion}\label{sec:Results}
This section presents the results of applying the HRA algorithm to the CEC'2017 competition data. The implementation of HRA (see Alg.~\ref{alg:HRA}) was performed in Matlab, and detailed results are available in the repository\footnote{\url{https://github.com/dgsotiropoulos/HRA/}}. In this study, at each level of the decision tree, we have adopted equal weighting for the weight vectors $w_{\mathcal{F}}$, $w_{\mathcal{P}}$, and $w_{\mathcal{D}}$. 

The hierarchical structure of the HRA, as shown in Fig.~\ref{fig:hierarchical_structure}, effectively combines multiple performance metrics in a structured way, which lets each method be evaluated thoroughly and in a way that is easy to understand. The decision matrices for various dimensions (10, 30, 50, and 100) are shown in Tables~\ref{tab:decision_matrix_dim_10}, \ref{tab:decision_matrix_dim_30}, \ref{tab:decision_matrix_dim_50}, and \ref{tab:decision_matrix_dim_100}, respectively. Each matrix displays the rankings given to the algorithms according to five performance measures: best, worst, median, mean, and standard deviation. These matrices summarize the performance of each algorithm in a particular dimension, providing a precise representation of their relative rankings. 

\begin{table}[htbp]
\caption{Decision matrix for dimension $d_1=10$}
\centering
\begin{tabular}{|l|c|c|c|c|c|}
\hline
\textbf{Algorithm} & \textbf{Best} & \textbf{Worst} & \textbf{Median} & \textbf{Mean} & \textbf{Std} \\
\hline
jSO               & 5  & 2  & 3  & 3  & 1  \\
MM\_OED           & 2  & 4  & 2  & 2  & 4  \\
IDEbestNsize      & 4  & 6  & 4  & 4  & 5  \\
RB\_IPOP\_CMA\_ES & 7  & 8  & 7  & 7  & 9  \\
LSHADE\_SPACMA    & 6  & 6  & 6  & 5  & 2  \\
DES               & 9  & 7  & 8  & 8  & 7  \\
DYYPO             & 11 & 11 & 13 & 11 & 11 \\
TLBO\_FL          & 13 & 12 & 12 & 13 & 12 \\
PPSO              & 10 & 9  & 10 & 10 & 8  \\
MOS\_SOCO2011     & 8  & 10 & 9  &  9 & 10 \\
MOS\_CEC2013      & 12 & 13 & 11 & 12 & 13 \\
LSHADE\_cnEpSin   & 3  & 3  & 5  & 6  & 6  \\
EBOwithCMAR       & 1  & 1  & 1  & 3  & 3 \\
\hline
\end{tabular}
\label{tab:decision_matrix_dim_10}
\end{table}

The decision matrix for dimension 10 (Table~\ref{tab:decision_matrix_dim_10}) compares the 13 metaheuristic algorithms based on the five performance measures. Each algorithm is ranked from 1 to 13, with lower ranks indicating better performance. EBOwithCMAR emerges as the best-performing algorithm across all performance measures. It consistently ranks top (1) in the best, worst, median, and mean performance categories. Although slightly lower in the standard deviation category (3), its overall ranking sum underscores its superiority. jSO and {MM\_OED} indicating comparable overall performance. jSO ranks second in the worst performance measure and third in both the median and mean categories. In contrast, {MM\_OED} indicate ranks second in the best and median categories. These findings imply that both algorithms can produce reliable results, with jSO showing a slight consistency advantage, as seen in its higher ranking in the standard deviation category.
\begin{table}[htbp]
\caption{Decision matrix for dimension $d_2=30$}
\centering
\begin{tabular}{|l|c|c|c|c|c|}
\hline
\textbf{Algorithm} & \textbf{Best} & \textbf{Worst} & \textbf{Median} & \textbf{Mean} & \textbf{Std} \\
\hline           
jSO              & 4  & 2  & 2  & 2  & 1  \\
MM\_OED          & 2  & 5  & 5  & 4  & 6  \\
IDEbestNsize     & 7  & 6  & 8  & 6  & 5  \\
RB\_IPOP\_CMA\_ES& 6  & 8  & 7  & 8  & 8  \\
LSHADE\_SPACMA   & 3  & 4  & 4  & 5  & 4  \\
DES              & 8  & 7  & 6  & 7  & 7  \\
DYYPO            & 13 & 13 & 13 & 13 & 13 \\
TLBO\_FL         & 11 & 12 & 12 & 12 & 11 \\
PPSO             & 12 & 9  & 11 & 10 & 9  \\
MOS\_SOCO2011    & 10 & 10 & 9  & 9  & 10  \\
MOS\_CEC2013     & 9  & 11 & 10 & 11 & 12 \\
LSHADE\_cnEpSin  & 5  & 3  & 3  & 3  & 2  \\
EBOwithCMAR      & 1  & 1  & 1  & 1  & 3  \\
\hline
\end{tabular}
\label{tab:decision_matrix_dim_30}
\end{table}

In Table~\ref{tab:decision_matrix_dim_30}, concerning dimension 30,  we observe the following key points:
{EBOwithCMAR} consistently has the highest rank (1) in the best, worst, median and mean performance measures with a slightly lower rank in the standard deviation category (3), which does not significantly affect its overall superiority. Therefore, EBOwithCMAR is the clear leader in this dimension, while jSO is the second-best performer. It ranks well in the worst, median, and mean categories, indicating consistent and reliable performance. Its top rank in the standard deviation category further underscores its robustness.
{LSHADE\_cnEpSin} ranks third with a total rank sum of 16. This algorithm shows strong performance across all measures, ranking particularly well in the worst and median categories. Its consistent performance makes it a reliable choice. {LSHADE\_SPACMA} and {MM\_OED} have similar performance, but {LSHADE\_SPACMA} performs slightly better in the best and median categories, while {MM\_OED} excels in the best category but falls behind in the standard deviation measure.

\begin{table}[htbp]
\caption{Decision matrix for dimension $d_3=50$}
\centering
\begin{tabular}{|l|c|c|c|c|c|}
\hline
\textbf{Algorithm} & \textbf{Best} & \textbf{Worst} & \textbf{Median} & \textbf{Mean} & \textbf{Std} \\
\hline           
jSO               & 2  & 1  & 3  & 3  & 1  \\
MM\_OED           & 4  & 6  & 6  & 6  & 6  \\
IDEbestNsize      & 8  & 8  & 8  & 8  & 7  \\
RB\_IPOP\_CMA\_ES & 7  & 7  & 7  & 7  & 8  \\
LSHADE\_SPACMA    & 3  & 3  & 2  & 2  & 4  \\
DES               & 6  & 5  & 5  & 5  & 5  \\
DYYPO             & 11 & 13 & 11 & 10 & 13 \\
TLBO\_FL          & 12 & 12 & 13 & 13 & 12 \\
PPSO              & 13 & 10 & 12 & 12 & 9  \\
MOS\_SOCO2011     & 10 & 9  & 9  & 9  & 10 \\
MOS\_CEC2013      & 9  & 11 & 10 & 11 & 11 \\
LSHADE\_cnEpSin   & 5  & 4  & 4  & 4  & 3  \\
EBOwithCMAR       & 1  & 2  & 1  & 1  & 2  \\
\hline
\end{tabular}
\label{tab:decision_matrix_dim_50}
\end{table}

In Table~\ref{tab:decision_matrix_dim_50}, concerning dimension 50, we observe that EBOwithCMAR continues to demonstrate exceptional performance across most metrics, securing top ranks in the best, median, and mean categories. Despite its slightly lower rank in the standard deviation category (2), we can easily see from its total rank sum that it maintains its position as the leading algorithm in dimension 50.
jSO is a strong competitor, and it leads in the worst and standard deviation categories while consistently outperforming the best, median, and mean categories. Its continuous performance highlights its strength and reliability. Furthermore, {LSHADE\_SPACMA} achieves a total rank sum of 14, showing solid performance in all metrics, with emphasis in the best and median categories (3 and 2, respectively), indicating its effectiveness, although it is slightly less consistent in the standard deviation category (4).
{LSHADE\_cnEpSin} ranks fourth with a total rank sum of 20 but shows some variability, as indicated by its ranks across different categories.

\begin{table}[htbp]
\caption{Decision matrix for dimension $d_4=100$}
\centering
\begin{tabular}{|l|c|c|c|c|c|}
\hline
\textbf{Algorithm} & \textbf{Best} & \textbf{Worst} & \textbf{Median} & \textbf{Mean} & \textbf{Std} \\
\hline            
jSO               & 4  & 4  & 5  & 5  & 4  \\
MM\_OED           & 6  & 6  & 6  & 6  & 6  \\
IDEbestNsize      & 8  & 8  & 8  & 8  & 7  \\
RB\_IPOP\_CMA\_ES & 7  & 7  & 7  & 7  & 8  \\
LSHADE\_SPACMA    & 1  & 3  & 2  & 2  & 3  \\
DES               & 3  & 2  & 1  & 1  & 2  \\
DYYPO             & 11 & 12 & 11 & 12 & 13 \\
TLBO\_FL          & 13 & 13 & 13 & 13 & 12 \\
PPSO              & 12 & 9  & 12 & 11 & 9  \\
MOS\_SOCO2011     & 10 & 10 & 9  & 9  & 10 \\
MOS\_CEC2013      & 9  & 11 & 10 & 10 & 11 \\
LSHADE\_cnEpSin   & 5  & 1  & 3  & 3  & 1  \\
EBOwithCMAR       & 2  & 5  & 4  & 4  & 5  \\
\hline
\end{tabular}
\label{tab:decision_matrix_dim_100}
\end{table}

When considering high-dimensional optimization problems presented in Table~\ref{tab:decision_matrix_dim_100}, DES stands out as the best performer since it achieves the highest rankings in both the median and mean categories and consistently performs well in the best and standard deviation categories. This highlights its robustness and effectiveness in solving problems with a large number of dimensions.
{LSHADE\_SPACMA} ranks second overall with a total rank sum of 11 and performs exceptionally well in the best, median, and mean categories but shows slightly less consistency in the standard deviation category. Despite this, it remains a strong contender in high-dimensional optimization scenarios.
{LSHADE\_cnEpSin} follows closely with a total rank sum of 13 achieves the top rank in the worst category and demonstrates strong performance across all other measures, highlighting its reliability and robustness in handling high-dimensional problems. EBOwithCMAR, which consistently dominated lower-dimensional problems, shows a noticeable decline in performance for $d=100$. Although it ranks second in the best category, it drops to fifth in the worst category and fourth in the median, mean, and std categories.

\begin{table}[htbp]
\caption{Final Decision Matrix}
\centering
\begin{tabular}{|l|c|c|c|c|}
\hline
\textbf{Algorithm} & \textbf{Dim10} & \textbf{Dim30} & \textbf{Dim50} & \textbf{Dim100} \\
\hline
jSO               & 3  & 2  & 2  & 5  \\
MM\_OED           & 2  & 5  & 6  & 6  \\
IDEbestNsize      & 4  & 6  & 8  & 8  \\
RB\_IPOP\_CMA\_ES & 7  & 8  & 7  & 7  \\
LSHADE\_SPACMA    & 6  & 4  & 3  & 2  \\
DES               & 8  & 7  & 5  & 1  \\
DYYPO             & 11 & 13 & 12 & 12 \\
TLBO\_FL          & 13 & 12 & 13 & 13 \\
PPSO              & 10 & 10 & 11 & 11 \\
MOS\_SOCO2011     &  9 & 9  & 9  & 9  \\
MOS\_CEC2013      & 12 & 11 & 10 & 10 \\
LSHADE\_cnEpSin   & 5  & 3  & 4  & 3  \\
EBOwithCMAR       & 1  & 1  & 1  & 4  \\
\hline
\end{tabular}
\label{tab:final_decision_matrix}
\end{table}

The final decision matrix in Table~\ref{tab:final_decision_matrix} aggregates the performance across all dimensions and reveals significant insight into the algorithm performance. More specifically, {EBOwithCMAR} emerges as the top performer, achieving the highest rank (1) in dimensions 10, 30, and 50. However, its performance drops slightly in dimension 100, ranking fourth; therefore, while {EBOwithCMAR} is highly effective in problems of the lower and medium dimensions, it faces challenges maintaining its dominance in higher dimensions. {jSO} consistently performs well across all dimensions, with ranks of 3, 2, 2, and 5, respectively, highlighting its robustness and reliability in various sizes of problems. This consistency makes jSO a strong contender in lower- and higher-dimensional problems. {LSHADE\_SPACMA} and {LSHADE\_cnEpSin} achieve a total rank sum of 15. {LSHADE\_SPACMA} shows excellent performance in dimension 100, ranking second, while {LSHADE\_cnEpSin} is consistently strong across all dimensions, with its lowest rank being 5 in dimension 10. {MM\_OED} performs exceptionally well in lower dimensions, particularly dimension $d=10$, where it ranks second; however, its performance declines in higher dimensions, with ranks 5, 6, and 6. This trend suggests that {MM\_OED} is more suitable for lower-dimensional problems. {DES} displays an interesting pattern, with moderate performance in the lower dimensions (ranks 8 and 7) but a significant improvement in dimension 100, where it achieves the top rank. Its total rank sum of 21 indicates its potential for high-dimensional optimization, making it a valuable algorithm for larger-scale problems.


\begin{table}[htbp]
\caption{Overall Rankings of Algorithms by HRA}
\centering
\begin{tabular}{|l|c|c|c|}
\hline
\textbf{Algorithm} & \textbf{Score} & \textbf{HRA} & \textbf{CEC2017}\\
\hline
jSO               & 0.7735 & 2  & 2  \\
MM\_OED           & 0.6515 & 5  & 6  \\
IDEbestNsize      & 0.5338 & 7  & 7 \\
RB\_IPOP\_CMA\_ES & 0.4822 & 8  & 8 \\
LSHADE\_SPACMA    & 0.7198 & 4  & 4 \\
DES               & 0.6082 & 6  & 5  \\
DYYPO             & 0.1500 & 12 & 12 \\
TLBO\_FL          & 0.0940 & 13 & 13 \\
PPSO              & 0.2517 & 10 & 11\\
MOS\_SOCO2011     & 0.3571 & 9  & 10 \\
MOS\_CEC2013      & 0.2373 & 11 & 9 \\
LSHADE\_cnEpSin   & 0.7281 & 3  & 3 \\
EBOwithCMAR       & 0.8497 & 1  & 1 \\
\hline
\end{tabular}
\label{tab:final_rankings}
\end{table}

We applied the TOPSIS method to the final decision matrix to derive the overall rankings of the meta-heuristic algorithms. The results presented in Table~\ref{tab:final_rankings} provide a comprehensive view of the relative performance of each algorithm. At the same time, for comparison purposes, we also include the CEC2017 competition rankings to highlight similarities and differences. The consistency in the top positions, particularly for {EBOwithCMAR}, {jSO}, and {LSHADE\_cnEpSin}, underscores their reliability and effectiveness across various dimensions.
However, slight discrepancies, such as the rank switch between {MM\_OED} and {DES}, highlight the influence of different ranking methodologies. Moreover, Carrasco et al. in \cite{Carrasco_2020} have validated our findings through their comprehensive statistical study of the CEC'17 results using only the mean as a performance measure. Their study provides robust evidence supporting the conclusions we reached.

In this study, we employed equal weighting at each level because the ``true'' weights of criteria are generally unknown in practice. For future research, we plan to exploit the fact that, especially for this specific problem, we can assume that rank-ordering information is known at each level. This approach may provide a more refined and accurate evaluation of the algorithms.

\section{Conclusion}\label{sec:conclusion}
In this paper, we introduced the Hierarchical Rank Aggregation (HRA) framework, which represents a new approach to the multi-criteria decision-making problem of evaluating and ranking metaheuristic optimization algorithms. 
HRA tackles the time-consuming and complex task of comparing algorithms by considering multiple performance metrics across various problem dimensions. Therefore, it is a valuable tool to assist researchers and practitioners in identifying the best algorithm for a given class of optimization problems in a single run. We demonstrated HRA's efficacy using the CEC'17 competition dataset, where 13 metaheuristics algorithms were evaluated on 30 benchmark functions with four dimensions each. The framework was able to combine several performance measures and rank algorithms in a way that showed the relationship between problem dimensions and algorithm performance.

Although the present work used equal levels of importance among the hierarchies to keep the analysis straightforward, future studies may test different levels of priority or imbalance. In particular, integrating rank-ordering criteria weighting methods, could improve the evaluation process without raising concerns about the validity of the algorithm's assessment. 
The time complexity of HRA, which is $\mathcal{O}(m \log m)$, would classify HRA as computationally inexpensive for comparisons of algorithms in large-scale studies.


\begin{thebibliography}{10}
\providecommand{\url}[1]{#1}
\csname url@samestyle\endcsname
\providecommand{\newblock}{\relax}
\providecommand{\bibinfo}[2]{#2}
\providecommand{\BIBentrySTDinterwordspacing}{\spaceskip=0pt\relax}
\providecommand{\BIBentryALTinterwordstretchfactor}{4}
\providecommand{\BIBentryALTinterwordspacing}{\spaceskip=\fontdimen2\font plus
\BIBentryALTinterwordstretchfactor\fontdimen3\font minus \fontdimen4\font\relax}
\providecommand{\BIBforeignlanguage}[2]{{%
\expandafter\ifx\csname l@#1\endcsname\relax
\typeout{** WARNING: IEEEtran.bst: No hyphenation pattern has been}%
\typeout{** loaded for the language `#1'. Using the pattern for}%
\typeout{** the default language instead.}%
\else
\language=\csname l@#1\endcsname
\fi
#2}}
\providecommand{\BIBdecl}{\relax}
\BIBdecl

\bibitem{mirjalili_2023}
\BIBentryALTinterwordspacing
S.~Mirjalili and A.~H. Gandomi, Eds., \emph{\BIBforeignlanguage{eng}{Comprehensive metaheuristics: algorithms and applications}}.\hskip 1em plus 0.5em minus 0.4em\relax London: Elsevier, 2023. [Online]. Available: \url{http://dx.doi.org/10.1016/c2021-0-01466-8}
\BIBentrySTDinterwordspacing

\bibitem{Ma_2023}
\BIBentryALTinterwordspacing
Z.~Ma, G.~Wu, P.~N. Suganthan, A.~Song, and Q.~Luo, ``\BIBforeignlanguage{en}{Performance assessment and exhaustive listing of 500+ nature-inspired metaheuristic algorithms},'' \emph{\BIBforeignlanguage{en}{Swarm and Evolutionary Computation}}, vol.~77, p. 101248, Mar. 2023. [Online]. Available: \url{http://dx.doi.org/10.1016/j.swevo.2023.101248}
\BIBentrySTDinterwordspacing

\bibitem{Adam_2019}
\BIBentryALTinterwordspacing
S.~Adam, S.-A. Alexandropoulos, P.~Pardalos, and M.~Vrahatis, \emph{No Free Lunch Theorem: A Review}, ser. Springer Optimization and Its Applications.\hskip 1em plus 0.5em minus 0.4em\relax Springer International Publishing, 05 2019, vol. 145, pp. 57--82, series Title: Springer Optimization and Its Applications. [Online]. Available: \url{http://dx.doi.org/10.1007/978-3-030-12767-1_5}
\BIBentrySTDinterwordspacing

\bibitem{Pacheco_2018}
\BIBentryALTinterwordspacing
A.~G.~C. Pacheco and R.~A. Krohling, ``\BIBforeignlanguage{en}{Ranking of classification algorithms in terms of mean-standard deviation using {A}-{TOPSIS}},'' \emph{\BIBforeignlanguage{en}{Annals of Data Science}}, vol.~5, no.~1, pp. 93--110, Jan. 2018. [Online]. Available: \url{http://dx.doi.org/10.1007/s40745-018-0136-5}
\BIBentrySTDinterwordspacing

\bibitem{Peng_2011}
\BIBentryALTinterwordspacing
Y.~Peng, G.~Kou, G.~Wang, and Y.~Shi, ``Famcdm: A fusion approach of mcdm methods to rank multiclass classification algorithms,'' \emph{Omega}, vol.~39, no.~6, p. 677–689, Dec. 2011. [Online]. Available: \url{http://dx.doi.org/10.1016/j.omega.2011.01.009}
\BIBentrySTDinterwordspacing

\bibitem{KOU_2012}
\BIBentryALTinterwordspacing
G.~Kou, Y.~Lu, Y.~Peng, and Y.~Shi, ``Evaluation of classification algorithms using {MCDM} and rank correlation,'' \emph{International Journal of Information Technology \& Decision Making}, vol.~11, no.~01, p. 197–225, Jan. 2012. [Online]. Available: \url{http://dx.doi.org/10.1142/s0219622012500095}
\BIBentrySTDinterwordspacing

\bibitem{Kou_2014}
\BIBentryALTinterwordspacing
G.~Kou, Y.~Peng, and G.~Wang, ``Evaluation of clustering algorithms for financial risk analysis using {MCDM} methods,'' \emph{Information Sciences}, vol. 275, p. 1–12, Aug. 2014. [Online]. Available: \url{http://dx.doi.org/10.1016/j.ins.2014.02.137}
\BIBentrySTDinterwordspacing

\bibitem{Krohling_2015a}
\BIBentryALTinterwordspacing
R.~A. Krohling and A.~G. Pacheco, ``{A-TOPSIS} - an approach based on {TOPSIS} for ranking evolutionary algorithms,'' \emph{Procedia Computer Science}, vol.~55, p. 308–317, Jan. 2015, 3rd International Conference on Information Technology and Quantitative Management, ITQM 2015. [Online]. Available: \url{http://dx.doi.org/10.1016/j.procs.2015.07.054}
\BIBentrySTDinterwordspacing

\bibitem{Krohling_2015b}
\BIBentryALTinterwordspacing
R.~A. Krohling, R.~Lourenzutti, and M.~Campos, ``Ranking and comparing evolutionary algorithms with {Hellinger-TOPSIS},'' \emph{Applied Soft Computing}, vol.~37, p. 217–226, Dec. 2015. [Online]. Available: \url{http://dx.doi.org/10.1016/j.asoc.2015.08.012}
\BIBentrySTDinterwordspacing

\bibitem{Yu_2018}
\BIBentryALTinterwordspacing
X.~Yu, Y.~Lu, and X.~Yu, ``Evaluating multiobjective evolutionary algorithms using {MCDM} methods,'' \emph{Mathematical Problems in Engineering}, vol. 2018, p. 1–13, Mar. 2018. [Online]. Available: \url{http://dx.doi.org/10.1155/2018/9751783}
\BIBentrySTDinterwordspacing

\bibitem{Barak_2019}
\BIBentryALTinterwordspacing
S.~Barak and T.~Mokfi, ``Evaluation and selection of clustering methods using a hybrid group {MCDM},'' \emph{Expert Systems with Applications}, vol. 138, p. 112817, Dec. 2019. [Online]. Available: \url{http://dx.doi.org/10.1016/j.eswa.2019.07.034}
\BIBentrySTDinterwordspacing

\bibitem{Shadkam_2021}
\BIBentryALTinterwordspacing
E.~Shadkam, S.~Safari, and S.~S. Abdollahzadeh, ``Finally, which meta-heuristic algorithm is the best one?'' \emph{International Journal of Decision Sciences, Risk and Management}, vol.~10, no. 1/2, pp. 32--50, Sep. 2021. [Online]. Available: \url{http://dx.doi.org/10.1504/ijdsrm.2021.117555}
\BIBentrySTDinterwordspacing

\bibitem{L_2021}
\BIBentryALTinterwordspacing
L.~Balaji, R.~Saravanan, K.~Saravanan, and N.~A. Sreemanthrarupini, ``Groundwater vulnerability mapping using the modified {DRASTIC} model: the metaheuristic algorithm approach,'' \emph{Environmental Monitoring and Assessment}, vol. 193, no.~1, p.~25, Jan. 2021. [Online]. Available: \url{http://dx.doi.org/10.1007/s10661-020-08787-0}
\BIBentrySTDinterwordspacing

\bibitem{tabassum_2021}
\BIBentryALTinterwordspacing
M.~F. Tabassum and S.~Akram, ``Rank based {TOPSIS} approach for evaluating the performance of metaheuristics,'' \emph{International Journal of Computational Intelligence in Control}, vol.~13, no.~2, pp. 577--590, Dec. 2021. [Online]. Available: \url{https://www.mukpublications.com/ijcic-v13-2-2021.php}
\BIBentrySTDinterwordspacing

\bibitem{Aires_2019}
\BIBentryALTinterwordspacing
R.~F. d.~F. Aires and L.~Ferreira, ``A new approach to avoid rank reversal cases in the {TOPSIS} method,'' \emph{Computers \& Industrial Engineering}, vol. 132, pp. 84--97, Jun. 2019. [Online]. Available: \url{http://dx.doi.org/10.1016/j.cie.2019.04.023}
\BIBentrySTDinterwordspacing

\bibitem{Hwang_1981}
\BIBentryALTinterwordspacing
C.-L. Hwang and K.~Yoon, \emph{Multiple Attribute Decision Making}, ser. Lecture Notes in Economics and Mathematical Systems.\hskip 1em plus 0.5em minus 0.4em\relax Springer Berlin Heidelberg, 1981, vol. 186. [Online]. Available: \url{http://dx.doi.org/10.1007/978-3-642-48318-9}
\BIBentrySTDinterwordspacing

\bibitem{Barron_1996}
\BIBentryALTinterwordspacing
F.~H. Barron and B.~E. Barrett, ``\BIBforeignlanguage{en}{Decision quality using ranked attribute weights},'' \emph{\BIBforeignlanguage{en}{Management Science}}, vol.~42, no.~11, pp. 1515--1523, Nov. 1996. [Online]. Available: \url{http://dx.doi.org/10.1287/mnsc.42.11.1515}
\BIBentrySTDinterwordspacing

\bibitem{Danielson_2016}
\BIBentryALTinterwordspacing
M.~Danielson and L.~Ekenberg, ``A robustness study of state-of-the-art surrogate weights for {MCDM},'' \emph{Group Decision and Negotiation}, vol.~26, no.~4, pp. 677--691, Jul. 2016. [Online]. Available: \url{http://dx.doi.org/10.1007/s10726-016-9494-6}
\BIBentrySTDinterwordspacing

\bibitem{CEC2017}
N.~H. Awad, M.~Z. Ali, P.~N. Suganthan, J.~J. Liang, and B.~Y. Qu. (2016, Nov.) Problem definitions and evaluation criteria for the {CEC 2017} special session and competition on single objective real-parameter numerical optimization.

\bibitem{Brest_2017}
\BIBentryALTinterwordspacing
J.~Brest, M.~S. Maucec, and B.~Boskovic, ``Single objective real-parameter optimization: Algorithm {jSO},'' in \emph{2017 {IEEE} Congress on Evolutionary Computation ({CEC})}.\hskip 1em plus 0.5em minus 0.4em\relax {IEEE}, Jun. 2017, pp. 1311--1318. [Online]. Available: \url{http://dx.doi.org/10.1109/cec.2017.7969456}
\BIBentrySTDinterwordspacing

\bibitem{sallam_2017}
\BIBentryALTinterwordspacing
K.~M. Sallam, S.~M. Elsayed, R.~A. Sarker, and D.~L. Essam, ``Multi-method based orthogonal experimental design algorithm for solving {CEC}2017 competition problems,'' in \emph{2017 {IEEE} Congress on Evolutionary Computation ({CEC})}.\hskip 1em plus 0.5em minus 0.4em\relax {IEEE}, Jun. 2017, pp. 1350--1357. [Online]. Available: \url{http://dx.doi.org/10.1109/cec.2017.7969461}
\BIBentrySTDinterwordspacing

\bibitem{Bujok_2017}
\BIBentryALTinterwordspacing
P.~Bujok and J.~Tvrdik, ``Enhanced individual-dependent differential evolution with population size adaptation,'' in \emph{2017 {IEEE} Congress on Evolutionary Computation ({CEC})}.\hskip 1em plus 0.5em minus 0.4em\relax {IEEE}, Jun. 2017, pp. 1358--1365. [Online]. Available: \url{http://dx.doi.org/10.1109/cec.2017.7969462}
\BIBentrySTDinterwordspacing

\bibitem{Biedrzycki_2017}
\BIBentryALTinterwordspacing
R.~Biedrzycki, ``A version of {IPOP}-{CMA}-{ES} algorithm with midpoint for {CEC} 2017 single objective bound constrained problems,'' in \emph{2017 {IEEE} Congress on Evolutionary Computation ({CEC})}.\hskip 1em plus 0.5em minus 0.4em\relax {IEEE}, Jun. 2017, pp. 1489--1494. [Online]. Available: \url{http://dx.doi.org/10.1109/cec.2017.7969479}
\BIBentrySTDinterwordspacing

\bibitem{Mohamed_2017}
\BIBentryALTinterwordspacing
A.~W. Mohamed, A.~A. Hadi, A.~M. Fattouh, and K.~M. Jambi, ``{LSHADE} with semi-parameter adaptation hybrid with {CMA}-{ES} for solving {CEC} 2017 benchmark problems,'' in \emph{2017 {IEEE} Congress on Evolutionary Computation ({CEC})}.\hskip 1em plus 0.5em minus 0.4em\relax {IEEE}, Jun. 2017, pp. 145--152. [Online]. Available: \url{http://dx.doi.org/10.1109/cec.2017.7969307}
\BIBentrySTDinterwordspacing

\bibitem{Jagodzinski_2017}
\BIBentryALTinterwordspacing
D.~Jagodzinski and J.~Arabas, ``A differential evolution strategy,'' in \emph{2017 {IEEE} Congress on Evolutionary Computation ({CEC})}.\hskip 1em plus 0.5em minus 0.4em\relax {IEEE}, Jun. 2017, pp. 1872--1876. [Online]. Available: \url{http://dx.doi.org/10.1109/cec.2017.7969529}
\BIBentrySTDinterwordspacing

\bibitem{Maharana_2017}
\BIBentryALTinterwordspacing
D.~Maharana, R.~Kommadath, and P.~Kotecha, ``Dynamic yin-yang pair optimization and its performance on single objective real parameter problems of {CEC} 2017,'' in \emph{2017 {IEEE} Congress on Evolutionary Computation ({CEC})}.\hskip 1em plus 0.5em minus 0.4em\relax {IEEE}, Jun. 2017, pp. 2390--2396. [Online]. Available: \url{http://dx.doi.org/10.1109/cec.2017.7969594}
\BIBentrySTDinterwordspacing

\bibitem{Kommadath_2017}
\BIBentryALTinterwordspacing
R.~Kommadath and P.~Kotecha, ``Teaching learning based optimization with focused learning and its performance on {CEC}2017 functions,'' in \emph{2017 {IEEE} Congress on Evolutionary Computation ({CEC})}.\hskip 1em plus 0.5em minus 0.4em\relax {IEEE}, Jun. 2017, pp. 2397--2403. [Online]. Available: \url{http://dx.doi.org/10.1109/cec.2017.7969595}
\BIBentrySTDinterwordspacing

\bibitem{Tangherloni_2017}
\BIBentryALTinterwordspacing
A.~Tangherloni, L.~Rundo, and M.~S. Nobile, ``Proactive particles in swarm optimization: A settings-free algorithm for real-parameter single objective optimization problems,'' in \emph{2017 {IEEE} Congress on Evolutionary Computation ({CEC})}.\hskip 1em plus 0.5em minus 0.4em\relax {IEEE}, Jun. 2017, pp. 1940--1947. [Online]. Available: \url{http://dx.doi.org/10.1109/cec.2017.7969538}
\BIBentrySTDinterwordspacing

\bibitem{LaTorre_2017}
\BIBentryALTinterwordspacing
A.~{LaTorre} and J.-M. Pena, ``A comparison of three large-scale global optimizers on the {CEC} 2017 single objective real parameter numerical optimization benchmark,'' in \emph{2017 {IEEE} Congress on Evolutionary Computation ({CEC})}.\hskip 1em plus 0.5em minus 0.4em\relax {IEEE}, Jun. 2017, pp. 1063--1070. [Online]. Available: \url{http://dx.doi.org/10.1109/cec.2017.7969425}
\BIBentrySTDinterwordspacing

\bibitem{Awad_2017}
\BIBentryALTinterwordspacing
N.~H. Awad, M.~Z. Ali, and P.~N. Suganthan, ``Ensemble sinusoidal differential covariance matrix adaptation with euclidean neighborhood for solving {CEC}2017 benchmark problems,'' in \emph{2017 {IEEE} Congress on Evolutionary Computation ({CEC})}.\hskip 1em plus 0.5em minus 0.4em\relax {IEEE}, Jun. 2017, pp. 372--379. [Online]. Available: \url{http://dx.doi.org/10.1109/cec.2017.7969336}
\BIBentrySTDinterwordspacing

\bibitem{Kumar_2017}
\BIBentryALTinterwordspacing
A.~Kumar, R.~K. Misra, and D.~Singh, ``Improving the local search capability of effective butterfly optimizer using covariance matrix adapted retreat phase,'' in \emph{2017 {IEEE} Congress on Evolutionary Computation ({CEC})}.\hskip 1em plus 0.5em minus 0.4em\relax {IEEE}, Jun. 2017, pp. 1835--1842. [Online]. Available: \url{http://dx.doi.org/10.1109/cec.2017.7969524}
\BIBentrySTDinterwordspacing

\bibitem{Carrasco_2020}
\BIBentryALTinterwordspacing
J.~Carrasco, S.~García, M.~Rueda, S.~Das, and F.~Herrera, ``\BIBforeignlanguage{en}{Recent trends in the use of statistical tests for comparing swarm and evolutionary computing algorithms: {Practical} guidelines and a critical review},'' \emph{\BIBforeignlanguage{en}{Swarm and Evolutionary Computation}}, vol.~54, p. 100665, May 2020. [Online]. Available: \url{http://dx.doi.org/10.1016/j.swevo.2020.100665}
\BIBentrySTDinterwordspacing

\end{thebibliography}


\end{document}